\documentclass[10pt, a4paper]{article}
\usepackage{lrec}
\usepackage{multibib}
\newcites{languageresource}{Language Resources}
\usepackage{graphicx}
\usepackage{tabularx}
\usepackage{soul}
\usepackage{subfig}
\usepackage{epstopdf}
\usepackage[utf8]{inputenc}

\usepackage{hyperref}
\usepackage{xstring}

\usepackage{color}

\title{A New Dataset for Natural Language Inference from Code-mixed Conversations}

\name{Simran Khanuja$^{a}$, Sandipan Dandapat$^{b}$,  Sunayana Sitaram$^{a}$, Monojit Choudhury$^{a}$}

\address{$^{a}$Microsoft Research India, $^{b}$Microsoft Corporation \\
         Bangalore, India, Hyderabad, India \\
         \{t-sikha, sadandap, sunayana.sitaram, monojitc\}@microsoft.com }

\abstract{
Natural Language Inference (NLI) is the task of inferring the logical relationship, typically entailment or contradiction, between a premise and hypothesis. Code-mixing is the use of more than one language in the same conversation or utterance, and is prevalent in multilingual communities all over the world. In this paper, we present the first dataset for code-mixed NLI, in which both the premises and hypotheses are in code-mixed Hindi-English. We use data from Hindi movies (Bollywood) as premises, and crowd-source hypotheses from Hindi-English bilinguals. We conduct a pilot annotation study and describe the final annotation protocol based on observations from the pilot. Currently, the data collected consists of 400 premises in the form of code-mixed conversation snippets and 2240 code-mixed hypotheses. We conduct an extensive analysis to infer the linguistic phenomena commonly observed in the dataset obtained. We evaluate the dataset using a standard mBERT-based pipeline for NLI and report results.\\ \newline  }

\begin{document}

\maketitleabstract

\section{Introduction}
Natural Language Inference (NLI) is a fundamental NLP task, not only because it has several practical applications, but also because it tests the language understanding abilities of machines beyond pattern recognition. NLI tasks usually involve inferring the logical relationship, such as entailment or contradiction, between a pair of sentences. In some cases, instead of a sentence, a document, paragraph or a dialogue snippet might be provided as the {\em premise}; the task then is to infer whether a given {\em hypothesis} is entailed in (or implied by) the premise. There are several monolingual NLI datasets available, with the most notable ones being included in the GLUE \cite{wang2018glue} and SuperGLUE \cite{wang2019superglue} benchmarks. There are also multilingual and crosslingual NLI datasets, such as XNLI \cite{conneau2018xnli}. These datasets have successfully spurred and facilitated research in this area. \\

In this paper, we introduce, for the first time, a new NLI dataset for {\em code-mixing}. Code-mixing or code-switching refers to the use of more than one language in a single conversation or utterance. It is prevalent in almost all multilingual societies across the world. Monolingual as well as multilingual NLP systems typically fail to handle code-mixed inputs. Therefore, recently, code-mixing has attained considerable attention from the speech and NLP communities. Consequently, there have been several shared tasks on language labeling, POS-tagging, and sentiment analysis of code-mixed text, and several datasets exist for these as well. Other speech and language processing tasks such as speech recognition, parsing, and question answering, have also been well researched upon. However, as far as we know, there exists no code-mixed dataset for any NLI task.  \\

The following reasons explain the motivation behind creating a code-mixed NLI dataset:
\begin{itemize}
    \item{NLI is an important requirement for chatbots and conversational agents, and since code-mixing is a spoken and conversational phenomenon, it is crucial that such systems understand code-mixing.}
    \item{Most NLI datasets, including monolingual datasets, are created using sentence pairs as the premise and hypothesis. Ours is one of the only datasets built on conversations as premises, which, we believe, facilitates improved consistency in dialogue agents. }
    \item{NLI helps indicate whether our models can truly understand code-mixing, as the task requires a deeper semantic understanding of language rather than reliance upon shallow heuristics.}
\end{itemize}
% Firstly, NLI is an important requirement for chatbots and conversational agents, and since code-mixing is a spoken and conversational phenomenon, it is crucial that such systems understand code-mixing. Secondly, although multilingual models have become popular recently, it is not clear whether zeroshot learning can solve code-mixed NLI. Most NLI datasets, including monolingual datasets, are created using sentence pairs as the premise and hypothesis. Ours is one of the only datasets built on conversations as premises. Lastly, NLI will tell us whether our models can truly understand code-mixing. \\

% 3 reasons for code-mixed NLI dataset: On the one hand, code-mixed NLI is an important end-application for chat bots because CM happens in chat - Alexa, Cortana etc. Also, it is not clear whether zeroshot learning can solve code-mixed NLI. NLI will tell us whether the machines truly understand CM. \\

To create the code-mixed NLI dataset, we use pre-existing code-mixed conversations from Hindi movies (\textit{Bollywood}) as premises, and ask crowd-workers to annotate the data with hypotheses that are either \textit{entailed in} or \textit{contradicted by} the premise. We follow this with a validation step where annotators are shown premises and hypotheses and are asked to validate whether the hypothesis is \textit{entailed in} or \textit{contradicted by} the corresponding premise. We conduct a pilot experiment and present its analysis with the final annotation scheme and a description of the data collection process. Currently our data consists of 400 premises with 2240 hypotheses in code-mixed \textit{Hindi-English}. \\

The rest of the paper is organized as follows. Section 2 introduces different NLI datasets and situates our work in their context. Section 3 describes the creation of the data for annotation. Section 4 describes the data annotation, including results from the pilot and the final annotation scheme. Section 5 presents an extensive analysis and a baseline evaluation. Section 6 concludes with a discussion of future work.\\
\begin{table*}[ht]
\begin{center}
\begin{tabular}{|l|l|}
      \hline
      Conversation & Translation\\
      \hline
      MRS.KAPOOR: Kitna old fashion hairstyle hai tumhara, & MRS KAPOOR: Your hairstyle is so old fashioned, \\
      new hair cut kyun nahin try karte .. Go to the Vidal & why don't you try a new hair cut .. Go to the Vidal \\  
      Sasoon salon tomorrow .. Aur thoda product use karo ..& Sasoon salon tomorrow .. And use some product .. \\
      You'll get some texture. & You'll get some texture.\\
      & \\
      MR.KAPOOR: Tumhari maa ko bahut pata hai, MBA kiya & MR.KAPOOR: Your mother knows a lot, she has \\
        hai usne hair styling mein. & done an MBA in hair styling.\\
     &  \\
      MRS.KAPOOR: Kaash kiya hota to tumhara kuch kar pati? & MRS.KAPOOR: I wish I had so that I could have done \\
      Kab se ke rahi hun, Soonawallas ki tarah hair transplant karva lo, & something about you? Been telling you for so long,\\
       already 55 ke lagte ho! & get a hair transplant like the Soonawallas, you already look \\
       &  like you are 55!\\
       & \\
      MR.KAPOOR: main 57 ka hun. & MR.KAPOOR: I am 57 years old.\\
      \hline
\end{tabular}
\caption{Example Conversation from the Bollywood data}
\label{Tab:EC}
\end{center}
\end{table*}
%end of pasted intro
\section{NLI Datasets}

% (Simran) 

NLI is a concept central to natural language understanding models. Most of the prominent datasets that are used to solve NLI problems involve learning textual entailment wherein we determine whether a hypothesis is entailed in or contradicts a textual document \cite{zhang2009we}. Even so, each dataset is severely limited in the reasoning it represents and cannot be generalised outside of its domain. \cite{bernardykind}  \\

\subsection{Types of NLI Datasets}
We briefly outline the prominent NLI datasets that have been well researched upon, to suitably place our contribution in context of the same.
\begin{itemize}

\item{The FraCaS test suite \cite{fracas1996using} consists of 346 manually curated premises followed by a \textit{Yes/No/Don't Know} question.}

\item{The RTE datasets \cite{dagan2005pascal} include naturally occurring data as premises and construct hypotheses based on them. All datasets have fewer than 1000 examples for training. A limitation of these datasets is that many examples assume world knowledge which is not explicitly labeled with each example.}

\item{The SNLI dataset \cite{bowman2015large} consists of 570k inference pairs created using crowd-sourcing on Amazon Mechanical Turk. The size of this dataset makes it conducive to be used for training deep learning models. Subjects are given the caption of an image and are asked to formulate a \textit{true} caption, a \textit{possible true} caption and a \textit{false} caption.}

\item{The Multi-Genre NLI corpus \cite{williams2017broad} is also a crowd-sourced collection of 433k sentence pairs annotated with entailment information. Although it is modeled on SNLI, it differs from it as it covers a variety of genres in both written and spoken English. XNLI \cite{conneau2018xnli} is a multilingual extension of MultiNLI wherein 5k (train) and 2.5k (dev) examples are translated into 14 languages.}

\item{The SICK (Sentences Involving Compositional Knowledge) \cite{marelli2014semeval} dataset consists of 9840 examples of inference patterns primarily to test distributional semantics. It is constructed by randomly selecting a subset of sentence pairs from two sources - the 8k ImageFlickr dataset and the SemEval2012 STS MSR-Video Description dataset.}

\item{The Dialogue NLI Corpus \cite{welleck2018dialogue} consists of pairs of sentences generated using the Persona-Chat dataset \cite{zhang2018personalizing}. Each human labeled triple is first associated to each persona sentence and then pairs of such triple; persona sentences are labeled as entailment, neutral or contradiction. The corpus consists of around 33k examples.}

\item{The Conversation Entailment \cite{zhang2010towards} dataset consists of 50 dialogues from the Switchboard corpus \cite{godfrey1992switchboard}. 15 volunteer annotators read the dialogues and manually created hypotheses to obtain a total of 1096 entailment annotated examples.}

\end{itemize}

While most of the datasets described above benefit information extraction and other textual analysis problems, they cannot be used to tackle inference in conversations, which is an important application today given the upsurge and importance of dialogue agents. \cite{bernardykind} make a strong case for the need of entailment datasets for dialogue data, highlighting that there has been no attempt towards building one so far. They point out several ways in which conversation entailment is different from textual entailment. Most importantly, each participant in the conversation adds more structure to the segment in his/her turn unlike textual entailment where one segment is stand-alone.\\

% % ****(can we make a reference to ruhh here where a lot of the conversations were code-mixed hence the need for code-mixed conversation nli?)****\\

Consider the example below from \cite{bernardykind}: \\

A. Mont Blanc is higher than 

B. Mt. Ararat? 

A. Yes. 

B. No, this is not correct. It is the other way around. 

A. Are you... 

B. Sure? Yes, I am. 

A. Ok, then \\

Further, with the exception of the XNLI dataset, all other NLI datasets are in English. This motivates us to use dialogue, or conversation, as a premise, and build hypotheses based on them for code-mixed language. Based on the approaches used for creating the datasets mentioned above, there are three main approaches that can be taken while creating a code-mixed NLI dataset. One approach is to translate an existing NLI dataset into a code-mixed language. Since there do not exist good Machine Translation systems for code-mixed languages, that can capture the nuances of the language necessary for an NLI dataset, this would need to be done manually to ensure high quality. Another approach is to synthesize code-mixed data artificially, using approaches such as \cite{pratapa2018language}. However, this cannot be done for a conversational dataset, and will not be natural enough to create good hypotheses. The third approach, which we take, is to use a naturally occurring source of conversational data as premises, and get the hypotheses manually annotated. \\

% hard problems, how hard, GLUE, harder datasets 
% Taxonomy of different kinds of NLI datasets – add all refs here for all other datasets 
% All can be converted to CM versions – generate but wont be natural 
% Motivate why we need to collect this data (vs. generate) 
 
% Caveat it is scripted 
 
%  NLI Datasets 

% A brief description of how each of the above datasets are constructed and what kind of examples they include :- \\

\section{Dataset Creation}

Code-mixing is primarily a spoken language phenomenon, so it is challenging to find naturally occurring code-mixed text on the web, or in standard monolingual corpora. Social Media and Instant Messaging data from multilingual users can be a source of code-mixed conversational data, but cannot be used due to privacy concerns. For this reason, we choose scripts of Hindi movies, also referred to as ``Bollywood" movies. Bollywood movies, from certain time periods and genres, contain varying amounts of code-mixing, as described in \cite{pratapa2017quantitative}. Although the movie data is not artificially generated, it is scripted, which makes it a less natural source of data than conversations between real people.

\subsection{Data Preparation}

The Bollywood data consists of scenes taken from 18 movies. The data is in Romanized form, so both Hindi and English parts of the conversation are written in the Roman script. Table \ref{Tab:EC} shows an example conversation from the Bollywood dataset. The data contains examples of both inter-sentential and intra-sentential code-mixing.\\

% \begin{table*}[ht]
% \begin{center}
% \begin{tabular}{|l|l|}
%       \hline
%       Conversation & Translation\\
%       \hline
%       MRS.KAPOOR: Kitna old fashion hairstyle hai tumhara, & MRS KAPOOR: Your hairstyle is so old fashioned, \\
%       new hair cut kyun nahin try karte .. Go to the Vidal & why don't you try a new hair cut .. Go to the Vidal \\  
%       Sasoon salon tomorrow .. Aur thoda product use karo ..& Sasoon salon tomorrow .. And use some product .. \\
%       You'll get some texture. & You'll get some texture.\\
%       & \\
%       MR.KAPOOR: Tumhari maa ko bahut pata hai, MBA kiya & MR.KAPOOR: Your mother knows a lot, she has \\
%         hai usne hair styling mein. & done an MBA in hair styling.\\
%      &  \\
%       MRS.KAPOOR: Kaash kiya hota to tumhara kuch kar pati? & MRS.KAPOOR: I wish I had so that I could have done \\
%       Kab se ke rahi hun, Soonawallas ki tarah hair transplant karva lo, & something about you? Been telling you for so long,\\
%       already 55 ke lagte ho! & get a hair transplant like the Soonawallas, you already look \\
%       &  like you are 55!\\
%       & \\
%       MR.KAPOOR: main 57 ka hun. & MR.KAPOOR: I am 57 years old.\\
%       \hline
% \end{tabular}
% \caption{Example Conversation from the Bollywood data}
% \label{Tab:EC}
% \end{center}
% \end{table*}

% Reason out the following paragraph : Rewrite

Based upon an initial manual inspection of the data, we make the following design choices :
\begin{itemize}
    \item{There are 1803 scenes in the 18 movie transcripts combined. We observe that a few scenes are monologues, reducing the problem from a conversational entailment to a textual one. Hence we use an initial filter of choosing scenes with greater than three number of turns.}
    \item{A number of scenes were in monolingual Hindi, this being a Bollywood movie dataset. Hence we calculate the \textit{Code Mixing Index (CMI)} \cite{gamback2014measuring} of each scene and choose scenes having a CMI greater than 20\%. After application of the above filters, we obtain 720 scenes.}
    \item{We choose not to transliterate the Romanized Hindi into the original Devanagari script. However, this can be done automatically using a transliteration system if desired.}
\end{itemize}

% There are 1803 scenes in the 18 moves in all. Not all the scenes contain code-mixing, so we use an initial filter based on the Code Mixing Index (CMI) \cite{gamback2014measuring}. We consider scenes with CMI greater than 20\%, and number of turns in the scene greater than 3 to remove monolingual and very short scenes, giving us 720 scenes. We further divide these scenes into three categories based on the number of tokens they contain to obtain 151 scenes that contain less than 55 tokens (Category 1), 252 scenes that contain less than 130 tokens (Category 2), and the rest containing more than 130 tokens (Category 3). We consider conversations from Categories 1 and 2 for annotation. Figures \ref{fig.1} and \ref{fig.2} show the distribution of conversations in Category 1 and 2, with number of tokens on the X axis and number of turns on the Y axis.\\

\begin{figure*}%
    \centering
    \subfloat[Category 1 distribution]{{\includegraphics[width=8cm, height=6cm]{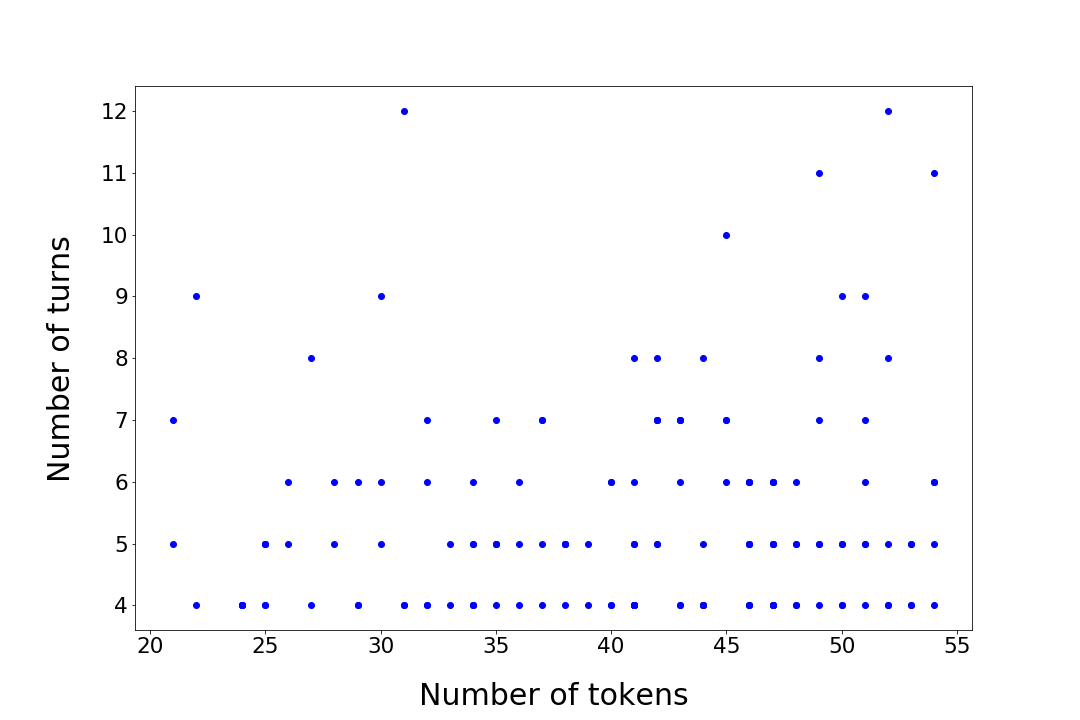} }}%
    \qquad
    \subfloat[Category 2 distribution]{{\includegraphics[width=8cm, height=6cm]{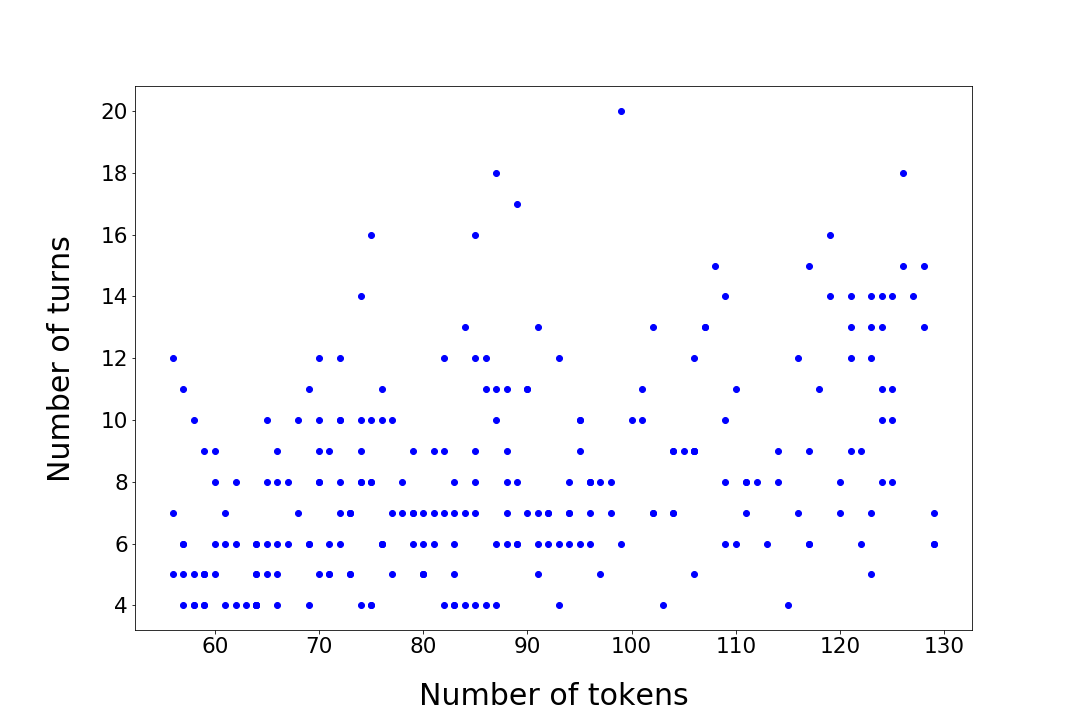} }}%
    \caption{Distribution of data in different categories}%
    \label{fig:example}%
\end{figure*}

%\begin{figure*}[h]
%\begin{center}
%%\fbox{\parbox{6cm}{
%%This is a figure with a caption.}}
%\includegraphics[scale=0.4]{rep_category_1.png} 
%\caption{Category 1 distribution}
%\label{fig.1}
%\end{center}
%\end{figure*}

%\begin{figure*}[h]
%\begin{center}
%%\fbox{\parbox{6cm}{
%%This is a figure with a caption.}}
%\includegraphics[scale=0.4]{rep_category_2.png} 
%\caption{Category 2 distribution}
%\label{fig.2}
%\end{center}
%\end{figure*}

% Further, we make the design choice not to transliterate the Romanized Hindi into the original Devanagari script, however, this can be done automatically using a transliteration system if desired.

\subsection{Task Paradigm}

The data annotation process involves the formulation of one or more true and false hypothesis, given a scene from the categories above as a premise. Subsequently, the NLI task is to classify whether the conversation entails the hypothesis or contradicts it, which we label true and false respectively. Note that the premises and the formulated hypotheses are in code-mixed \textit{Hindi-English}.\\

\section{Data Annotation}

% (Sandipan) 

\subsection{Initial annotation guidelines}
%Excel screenshot interface and instructions 
Our annotation scheme consists of two stages. In the first stage, we present conversations with a set of already created hypotheses (cf. Table~\ref{Tab:DH}) and ask the annotators to assign two labels to each hypothesis statement. The first is a \textit{true/false} label and the second is a \textit{good/fair/bad} label, judging the quality of the given hypothesis. Table~\ref{Tab:TL} shows details of the two different labels used in the annotation process.\\ \\
Annotators were also instructed to assign an {\it Irrelevant} label in case the generated hypothesis is not relevant to the conversation. In general, a hypothesis is considered as irrelevant when there is not enough topic and word overlap between the statement of the hypothesis and the conversation, especially when generating negative hypotheses, or when world knowledge is used to formulate the hypothesis, which cannot be inferred from the conversation.\\ \\
The first stage is conducted to fulfil two objectives: 
    \begin{itemize}
        \item{It acts as an initial filter to make sure that the annotators are well versed in both languages and have a good understanding of the task. If they fail to assign gold labels to more than 80 percent of the hypotheses, they will not be assigned the second stage of annotation.}
        \item{It serves to show annotators, the kind of hypotheses we are expecting will be generated from the conversations.}
    \end{itemize}
% In the initial annotation process, the annotators were presented with a conversation snippet as shown in Table \ref{Tab:EC}. The annotation was conducted in two stages. In the first stage, we presented conversations with a set of already created hypotheses (cf. Table~\ref{Tab:DH}) and asked the annotators to assign two labels to each hypothesis statement. Table~\ref{Tab:TL} shows details of the two different labels used in the annotation process. \\

\begin{table*}[ht]
\begin{center}
\begin{tabular}{ccc}
      \hline
      Label & Categories & Definition\\
      \hline
      Label1 & True & It can be inferred from the conversation (entailed)\\
      & False & It is contradictory to the conversation\\
      \hline
      & Good & An unambiguous statement which can clearly \\ & & be either inferred from the conversation or stands contradictory to it \\
      Label2 & Fair & Can be fairly inferred/contradicted from the conversation \\
      & & but lacks in either a good structure/is too long/is too abstract \\
      & & contains too many(or too few) words from the snippet \\
      & Bad & A statement which isn’t well-formed/ is too ambiguous \\
      & & or is verbatim from the conversation \\
      \hline
\end{tabular}
\caption{Types of labels}
\label{Tab:TL}
\end{center}
\end{table*}

% Annotators were also instructed to assign an {\it Irrelevant} label in case the generated hypothesis is not relevant to the conversation. In general, a hypothesis is considered as irrelevant when there is not enough topic and word overlap between the statement of the hypothesis and the conversation, especially when generating negative hypotheses, or when world knowledge is used to come up with the hypothesis which cannot be inferred from the conversation. Table~\ref{Tab:DH} shows some examples of different hypotheses created based on the conversation in Table~\ref{Tab:EC}. \\

%Hypotheses 
%Choose the correct example and NEED to discuss with examples
\begin{table*}[ht]
\begin{center}
\begin{tabular}{|l|l|l|}
      \hline
      Category&Hypothesis&Translation\\
      \hline
       True& Mr. Kapoor 57 years ke hai & Mr. Kapoor is 57 years old\\
False & Mrs. Kapoor ne hair styling mei MBA kiya hai & Mrs. Kapoor has done an MBA in hair styling\\
Bad & Mr. Kapoor will go to the Vidal Sasoon salon tomorrow&\\
Irrelevant & Mr. Kapoor was born in Delhi.&\\
Ambiguous  & Mrs. Kapoor ko hair styling ke baare mei bohot pata hai & Mrs. Kapoor knows a lot about hair styling\\
% Prior Knowledge &  &\\ 

      \hline

      \hline
\end{tabular}
\caption{Different kinds of hypotheses for the conversation snippet in Table \ref{Tab:EC}}
\label{Tab:DH}
\end{center}
\end{table*}

% ANDREA: Guys mein coffee bana rahi hoon , koi lega .. 

% MINAL : Main shower lene ja rahi hoon ! 
% FALAK: Guys aisey baat avoid karney se kuchh honey nahin wala .. 
% MINAL: Dekh yaar jo hogaya vo hogaya , they must have forgotten , hum bhi bhool jaatey hain yaar ... 
% FALAK: Itna easy nahin hai yaar ... 
% ANDREA: Cut kafi bada tha ... like this ... below his eye ... 
% MINAL: Look roz subah subah tense hone se kya faayda ! We'll go mad ... 
% ANDREA: I feel she is right ! Let us just forget the whole thing ya ! 
% FALAK: Vo bhool jaayengey ? 
% MINAL: Chaar din ho gaye - kuchh hua nahin na ... so why are we tense ? 
% FALAK: Chaar din ho gaye toh kya hansoon ! 
% MINAL: Haan ! 
% MINAL: Tu bhi hans ! 
% True 
% False 
% Bad 
% Irrelevant 
% Ambiguous  
% When people know the movie 

% The first stage was conducted to fulfil two objectives. Primarily, it acted as an initial filter to make sure that the annotators were well versed in both languages and have a good understanding of the task. If they failed to assign gold labels to more than 80 percent of the hypotheses, they would not be assigned the second stage of annotation. It also serves to show annotators the kind of hypotheses we are expecting will be generated from the conversations. \\

In the second stage, the annotators are given only the conversation snippet and are asked to come up with hypotheses which they think are \textit{entailed in} or \textit{contradicted by} the conversation. We provide annotators with a guideline containing worked out examples to make them familiar with the classification and help them generate good hypotheses. These hypotheses could be written in Hindi, English or both languages mixed in one sentence, as people often do in informal settings. Note that since the conversation contains Romanized Hindi, we ask the annotators to write Hindi in the Roman script. Romanized Hindi is not standardized, so we find variations of the same word across the Bollywood data. The annotators were asked to use spelling variants that they found in the snippets, or use the variants they are most familiar with.\\

% Note that the first stage of annotation helps to generate the training data for NLI task, while the second stage generates test data. \\ SUNAYANA - didn't quite get this
 
\subsection{Pilot experiments} 
%Examples of good/bad/true/false hypotheses – one passage 
%(add table here with one passage and good and bad hyps generated for that passage by multiple people) 
%What we learned 
%Long premises not good 
%Too many hypotheses problem 
%People know the movies 

In the pilot experiment, for the first stage, we use 2 conversations of different lengths (7 and 17 turns) having a set of carefully curated hypotheses (8 and 10 respectively). The task is to mark each hypothesis with Label1 (True/False/Irrelevant) and Label2 (Good/Fair/Bad). On an average, the number of correct labels is 88\%. \\ \\
For the second stage, we take 3 conversation snippets of different lengths (9, 12 and 13 turns) and ask the annotators to generate 4 hypotheses (2 True and 2 False) for each conversation. 7 different annotators conduct the task.\\

%TODO: stats about true/false from pilot
%TODO: results from the pilot

Our observations from the pilot are as follows: 
\begin{itemize}
    \item{Annotators do not prefer long premises as they need to go back and forth to validate the correctness of a statement. However, too short a premise also does not provide enough context for the annotators to come up with good hypotheses.}
    \item{Annotators face difficulty in producing a large number of hypotheses. The average amount of time required to produce a hypothesis increases non-linearly with the number of hypotheses expected from a conversation.}
    \item{A few annotators use prior knowledge about the topic (i.e. the movie is known to the annotator). This leads to the generation of bad hypotheses or incorrect labeling.}
\end{itemize}
% Annotators did not prefer long premises as they needed to go back and forth to validate the correctness of a statement, however, too short a premise also did not provide enough context for the annotators to come up with good hypotheses. Annotators also had difficulties in producing a large number of hypotheses. The average amount of time required to produce a hypotheses increased with the number of hypotheses expected from a conversation. We also observed that some annotators used prior knowledge about the topic (i.e. the movie is known to the annotator). This led to the generation of bad hypotheses or incorrect labeling. \\

% We have observed that the annotators often do not prefer longer premises as they need to go back and forth to validate the correctness of a statement. The annotators also have found difficulties in producing larger number of hypotheses. In addition, the average amount time required to produce each statement increases with the number of hypotheses expected from a conversation. We also have found that the annotators are sometime biased with their prior knowledge about the topic (i.e. the movie is known to the annotator). This often leads to the generation of bad hypotheses and/or incorrect labeling which are not evident from the given snippet (but annotators uses facts from the movie). 
%<I think we need some quantitative results for some of the above claims> 
 
\subsection{Final scheme and guidelines }
%Length of premise 
%Number of hyp 
%Curriculum/order 
%De-biasing 

Based on the observations from the pilot experiments, we make the following changes into the annotation process: 

\begin{itemize}

\item{\textbf{Length of the Premises}: We segregate the conversations into three categories based on the number of tokens they contain, to obtain 151 scenes that contain less than 55 tokens (Category 1), 252 scenes that contain less than 130 tokens (Category 2), and the rest containing more than 130 tokens (Category 3). We consider conversations from Categories 1 and 2 for annotation, based on the observation that annotators find it increasingly time-consuming to formulate hypotheses for very long conversations. Figures \ref{fig:example} (a) and (b) give a pictorial representation of conversations in Category 1 and 2, with \textit{number of tokens} on the X axis and \textit{number of turns} on the Y axis.}

\item{\textbf{Number of Hypotheses}: Depending on the length of the premises, the annotators are asked to generate different number of hypotheses. The required number of hypotheses is 2 (one True and one False) if a conversation is from Category 1 (between 20-55 tokens) and 4 (two each for True and False) if taken from Category 2 (between 55 and 130 tokens). However, the annotators have the option to generate additional hypotheses if they desire.}

\item{\textbf{De-biasing}: Bias in NLI datasets is well studied \cite{rudinger2017social} and can be attributed to annotators amplifying stereotypical characteristics of the conversation participants. In our case, there is additional bias due to the knowledge of the movie, which can be inferred from the names of some characters, and sometimes from the conversation. To handle the latter, we anonymize the names of the turn owners and replace them with generic tokens (``C1", ``C2" etc.). In this process, we only substitute the proper names from the conversation and not the kinship terms (\textit{Father, Mother, Bauji etc.}) or professions (\textit{Doctor, Receptionist, Police Officer etc.}). This helps reduce the familiarity of the conversation with a known movie which produces noise in the pilot study (cf. Section 4.2).}

\end{itemize}

\subsection{Final annotation process}

The final hypotheses generation process is as follows: 
\begin{itemize}
    \item{First, an annotator is shown the conversation after making the changes described above, and asked to formulate 2 or 4 hypotheses depending on the length of the conversation. Currently, we have 600 hypotheses created from 150 premises in Category 1 (length between 20-55 tokens) and another 1640 hypotheses created from 250 premises in Category 2 (length between 55-130 tokens).}
    \item{Subsequently, we conduct a validation step in which two annotators are shown 300 conversation snippets and corresponding hypotheses, and asked to mark the hypotheses ``True" (entailed),``False" (contradicted) or ``Irrelevant". The Inter-Annotator Agreement is 0.863, and the agreement of each annotator with the labels of the generated hypotheses is greater than 0.8, which shows that the data collected is of good quality.}
\end{itemize}

\section{Analysis and Evaluation}
On a deeper analysis of the hypotheses generated, we make the following observations:
\begin{itemize}
\item{\textbf{Sarcasm and Rhetorics}: Several examples require the model to interpret sarcasm in the conversation, to make a correct prediction. This is natural, given the premises are human conversations, and these help add complexity to the dataset. For example -\\ \\
PREMISE:\\
Mother: 5 saal baad saath-saath aaye ho .. janvaron ki tarah ladna zaroori hai ? \\
C0: Haan aapko toh main hi galat lagta hoon .. \\ \\
HYPOTHESIS: \\
Mother told C0 to quarrel like animals. (\textit{False})\\ \\ 
\textbf{\textit{Translated}}\\ \\
PREMISE:\\
Mother: Y'all have met after 5 years .. is it necessary to fight like animals? \\
C0: Yeah you always think I am wrong .. \\ \\
HYPOTHESIS: \\
Mother told C0 to quarrel like animals. (\textit{False})
}
\item{\textbf{Word Sense Disambiguation} : There exist several examples requiring the model to resolve the meaning of the word in context of its usage. For example, in the following, the word "saala" is used as an abusive term in the premise, but is taken to mean "brother in law" in the hypothesis -\\ \\ 
PREMISE:\\ 
C0: Ek lafz aur toh tera bheja baahar . 
\\C1: Accha ? Nikaal .. Himmat hai to nikal C1 ka bheja baahar ! 
\\C1: Maar ! 
\\C0: Dekh be C1 . Aakhiri baar keh raha hoon .. 
\\C1: Naqli Nawab saala .. \\ \\
HYPOTHESIS: 
\\ C0 is C1's brother in law. (\textit{False}) \\ \\
\textbf{\textit{Translated}}\\ \\
PREMISE:\\ 
C0: One more word and I will smack your head. 
\\C1: Really ? Hit .. If you have the strength, hit me ! 
\\C1: Hit ! 
\\C0: See C1 . I am telling you one last time .. 
\\C1: Fool  .. \\ \\
HYPOTHESIS: 
\\ C0 is C1's brother in law. (\textit{False})
}
\item{\textbf{Inter-dependent Inference} : Several premises are such that each utterance is highly contextual, requiring knowledge of the speakers of the past few utterances as well. Hypotheses thus generated pick facts from several utterances at once. For example - \\ \\
PREMISE:\\
C0: Kaun se school mein tha ? \\C1: Bishop Cotton . \\C0: Kahan hai ? \\C1: Shimla ...\\ \\
HYPOTHESIS:\\
Bishop Cotton School Manali mein hai. (\textit{False}) \\ \\
\textbf{\textit{Translated}}\\ \\
PREMISE:\\
C0: Which school were you in ? \\C1: Bishop Cotton . \\C0: Where is it ? \\C1: Shimla ... \\ \\
HYPOTHESIS:\\
Bishop Cotton School is in Manali. (\textit{False})
}
\item{\textbf{Domain Generality} : We also observe that this being a movie dataset, we obtain premise-hypothesis pairs across several domains. There even exist pairs with dialect differences as shown below :- \\ \\
PREMISE:\\ C0: Chhorey tanne manaa karya tha na jaane se ? \\C1: Koi milne aaya hai . \\C0: Kaun ? \\C0: Kaun sa ? \\C1: Boli thaare se kaam tha\\ \\
HYPOTHESIS:\\ C0 ne C1 ko jaane se mana kiya tha. (\textit{True}) \\ \\
\textbf{\textit{Translated}}\\ \\
PREMISE:\\ C0: Son, I had told you not to go right ? \\C1: Somebody had come to meet me . \\C0: Who ? \\C0: Who was it ? \\C1: She said she had some work for you\\ \\
HYPOTHESIS:\\ C0 had told C1 not to go. (\textit{True})}
% \item{Code-mixing pattern: We also observe that the annotators tend to generate hypotheses in a similar pattern as to how it is present in the premise. [cite] show that code-mixing is highly user-dependent and } 

\item{\textbf{Speaker Conflict}: We also observe examples wherein multiple parties hold different beliefs on a particular fact, hence inferring about the fact from the conversation becomes a difficult task. For example - \\ \\
PREMISE:\\
C0: Waise main bhi uski tarah chest hila sakta hun. \\
C1: Show . See .. Nobody can beat him. \\ \\
HYPOTHESIS:\\
C0 bhi uski tarah chest hila sakta hai. (\textit{False})\\ \\
\textbf{\textit{Translated}}\\ \\
PREMISE:\\
C0: Even I can move my chest like him. \\
C1: Show . See .. Nobody can beat him. \\ \\
HYPOTHESIS:\\
C0 can also move his chest like him. (\textit{False})
}
\begin{table*}
\centering
\begin{tabular}{lccccc}
\hline 
 \textbf{Model} & \textbf{RTE} & \textbf{SNLI} & \textbf{MNLI} & \textbf{QNLI} \\
 \hline
  $BERT_{BASE}$  & 66.4 & 90.4 & 86.7 & 90.5\\
  \hline
 \end{tabular}
\caption{\label{BERT-results} NLI results (Accuracy)}
\end{table*}

\begin{table*}
\centering
\begin{tabular}{lcc}
\hline 
 \textbf{Model} & \textbf{NLI En-Hi}  \\
 \hline
  $mBERT$  &  57.82 \\
  \hline
 \end{tabular}
\caption{\label{mBERT-results} NLI results (Accuracy)}
\end{table*}
\item{\textbf{Paraphrasing}: In a few examples, true hypotheses are paraphrases of what was said in the conversation. In some cases, they are a substring of the conversation, but in other cases, they are paraphrased using code-mixing, or a single language when the premise uses the other language. This is usually observed in longer conversations. An example wherein the hypothesis is picked verbatim from the conversation is shown below : \\ \\ 
PREMISE:\\
C0: Nahi Sir busy hain - voh nahi le saktey brief aapka ! \\
C1: Lekin subah toh unhone kaha tha ki ...\\ \\
HYPOTHESIS:\\
Sir busy hain. (\textit{True})\\ \\
\textbf{\textit{Translated}}\\ \\
PREMISE:\\
C0: No, Sir is busy - He cannot take your brief ! \\
C1: But in the morning he said that ...\\ \\
HYPOTHESIS:\\
Sir is busy. (\textit{True})
}

\item{\textbf{Negation}: True or False hypotheses were negations of what was said in the conversation. For example - \\ \\
PREMISE: \\
C0: Kahin bhi shuru ho jaati ho dance karna , shushma didi ki sagai hai ... relations mein hain humarey ... socha to karo ... \\
C1: Baaki ladkiyan bhi to kar rahi thi ...\\ \\
HYPOTHESIS:\\
Baaki ladkiyan dance nahi kar rahi hai. (\textit{False}) \\ \\
\textbf{\textit{Translated}}\\ \\
PREMISE: \\
C0: You start dancing anywhere, It's sushma's reception ... they are our relatives... think sometimes \\
C1: But the other girls were dancing as well ...\\ \\
HYPOTHESIS:\\
The other girls are not dancing. (\textit{False})
}

\item{\textbf{Swapping Roles}: We also observe cases wherein a false hypothesis is constructed by simply swapping for the speaker. For example - \\ \\
PREMISE:\\
C1: Jaan bhai ! Ab kya hoga ?
\\C0: Sab theek ho jaayega . Chup kar bus chup . Sab theek ho jaayega . Bank manager ko bol 10 karod cash chahiye kal subah \\ \\
HYPOTHESIS: \\C1 bol raha hai sab theek ho jaaega. (\textit{False}) \\ \\
\textbf{\textit{Translated}}\\ \\
PREMISE:\\
C1: Brother ! What will happen now ?
\\C0: Everything will be alright. Just be quiet. Everything will be alright. Tell the bank manager to arrange for 10 crore rupees by tomorrow morning. \\ \\
HYPOTHESIS: \\C1 says that everything will be alright. (\textit{False})

}

\item{\textbf{Numerical Hypotheses}: A few examples simply change a numeral in the premise to create a false hypothesis. For example - \\ \\
PREMISE:\\
C2: Kitne saal se kaam kar rahe ho clinic mein ? \\C1: 4 to ho gaye honge saab .. \\ \\
HYPOTHESIS:\\ C2 5 saal se clinic mein kaam karta hai. (\textit{False}) \\ \\
\textbf{\textit{Translated}}\\ \\
PREMISE:\\
C2: For how many years have you been working at the clinic ? \\C1: It must have been 4 years at the least, Sir .. \\ \\
HYPOTHESIS:\\ C2 has been working at the clinic for 5 years. (\textit{False})
}

\item{\textbf{Length of Premise} : We also observe that for longer premises, annotators usually pick out sentences verbatim from the conversation. In general, the quality of the hypotheses generated decreases as the premises become longer.}

\item{No hypotheses are found that are irrelevant or use world knowledge, or knowledge about the movies.}

\end{itemize}

On the basis of the above observations, we see that the dataset obtained is highly varying in complexity. Models that rely on shallow heuristics and learn statistical patterns from training data, which is the case with most neural models today \cite{mccoy2019right}, are expected to correctly predict examples involving \textit{Negation}, \textit{Numeral Changes}, \textit{Swapping Roles} or \textit{Paraphrasing}. However, they are hypothesized to fail in examples requiring deeper semantic knowledge, for instance, the examples involving \textit{Sarcasm}, \textit{Word Sense Disambiguation}, \textit{Inter-dependent Inference} or \textit{Speaker Conflict}.\\

With the recent upsurge of multilingual models, and claims that they can be used to solve code-mixed tasks as well, we evaluate the multilingual BERT model on our dataset. Previously, it has been shown to perform well on code-mixed POS tagging by \cite{pires2019multilingual}. Our results are as shown in Table \ref{mBERT-results}. We make use of the \textit{transformers} library\footnote{https://github.com/huggingface/transformers} for the experiment. We use the AdamW optimizer with a learning rate of 5e-5, epsilon of 1e-8, and a batch size of 16, as suggested by \cite{devlin2018bert}. We train for 5 epochs. We report the average result of training on 5 random seed values. Note that the dataset contains Hindi in Roman script while mBERT is trained on Hindi in Devanagari, and we report this number as a mere baseline.\\

To put our numbers in perspective, we have included accuracies achieved by the BERT base model, as shown in \cite{talman2018testing} and \cite{devlin2018bert}, in Table \ref{BERT-results} on standard monolingual NLI datasets. Note that these numbers are not directly comparable due to differences in language and corpus sizes. However, even standalone, the accuracy obtained by mBERT on our dataset clearly highlights the fact that this task is far from being solved.  \\

% \begin{table*}
% \centering
% \begin{tabular}{lccccc}
% \hline 
%  \textbf{Model} & \textbf{RTE} & \textbf{SNLI} & \textbf{MNLI} & \textbf{QNLI} \\
%  \hline
%   \[BERT_{BASE} \] & 66.4 & 90.4 & 86.7 & 90.5\\
%   \hline
%  \end{tabular}
% \caption{\label{BERT-results} NLI results (Accuracy)}
% \end{table*}

% \begin{table*}
% \centering
% \begin{tabular}{lcc}
% \hline 
%  \textbf{Model} & \textbf{NLI En-Hi}  \\
%  \hline
%   mBERT  &  57.82 \\
%   \hline
%  \end{tabular}
% \caption{\label{mBERT-results} NLI results (Accuracy)}
% \end{table*}

%What the data collection plan going forward is.

% \section{Summary}

% What has been done and what the plan is (update last minute) 
% Data release

\section{Conclusion and Future Work}

In this paper, we introduce a new dataset for code-mixed Natural Language Inference (NLI). Our dataset is unique due to the nature of the language used (code-mixed Hindi-English) and also because it is one of the few datasets created using conversations as premises. Solving the NLI task would help understand how well machines understand code-mixing. We also observe that multilingual models such as mBERT \cite{devlin2019bert} are not competent enough to solve this task, thus highlighting the need for models especially suited for the task at hand. In future work, we plan to experiment with neural and symbolic architectures for code-mixed NLI. One challenge in testing our data on models pre-trained on monolingual data is a script mismatch, as monolingual models tend to be trained on Devanagari, while our data contains Romanized Hindi with spelling variations.\\

Given the  nature of the data, we observe that this dataset can be scaled up to generate a plethora of such premise hypothesis pairs. Noting the dearth of conversation entailment datasets in monolingual settings as well, the same can be done to create monolingual datasets. This can be a major contribution to help solve conversation inference tasks which can show significant improvements in existing conversational agents.\\

Currently, our dataset consists of 400 premises with 2240 hypotheses, labeled for True and False only. We plan to continue the annotation process with more such transcripts. Further, we plan to further annotate the dataset for other linguistic phenomena, which may help to better solve the task. We plan to release the annotations we have crowd-sourced for research purposes and hope that it will spur research in the field of code-mixed NLI.\\

\section{Bibliographical References}\label{reference}
%\label{main:ref}

\bibliographystyle{lrec}
\bibliography{lrec2020W-xample-kc}

% \section{Language Resource References}
% \label{lr:ref}
% \bibliographystylelanguageresource{lrec}
% %\bibliographylanguageresource{languageresource}

\end{document}